\documentclass[10pt]{article}
\usepackage[left=2.2cm,right=2.2cm,top=2.5cm,bottom=2.5cm]{geometry}

\usepackage{booktabs}
\usepackage{amsmath}
\usepackage{amsfonts}
\usepackage{amsthm}

\usepackage{url}
\usepackage[authoryear,square]{natbib}
\usepackage{cite}
\usepackage{epstopdf}
\usepackage[titletoc,toc,title]{appendix}
\usepackage{authblk}
\usepackage{epstopdf}
\usepackage{epsfig}
\usepackage{hyperref}

\usepackage{numprint}
\npthousandsep{,}\npthousandthpartsep{}
\usepackage{graphics}
\usepackage{graphicx}
\usepackage{xcolor}
\usepackage{algorithm}
\usepackage{algpseudocode}
\usepackage{enumitem} 

\mathchardef\mhyphen="2D

\renewcommand{\cite}{\citep}

\usepackage[skip=0.5\baselineskip]{caption}

\newcommand\blfootnote[1]{%
  \begingroup
  \renewcommand\thefootnote{}\footnote{#1}%
  \addtocounter{footnote}{-1}%
  \endgroup
}

\begin{document}


\title{Iterative Delexicalization for Improved Spoken Language Understanding}

\author{Avik Ray \thanks{avik.r@samsung.com}}
\author{Yilin Shen \thanks{yilin.shen@samsung.com}}
\author{Hongxia Jin \thanks{hongxia.jin@samsung.com}}
\affil{{Samsung Research America} \authorcr
{Mountain View, CA, USA}}

\date{$15^{th}$ October $2019$}

\maketitle
\begin{abstract}
Recurrent neural network (RNN) based joint intent classification and slot tagging models have achieved tremendous success in recent years for building spoken language understanding and dialog systems. However, these models suffer from poor performance for slots which often encounter large semantic variability in slot values after deployment (e.g. message texts, partial movie/artist names). While greedy delexicalization of slots in the input utterance via substring matching can partly improve performance, it often produces incorrect input. Moreover, such techniques cannot delexicalize slots with out-of-vocabulary slot values not seen at training. In this paper, we propose a novel iterative delexicalization algorithm, which can accurately delexicalize the input, even with out-of-vocabulary slot values. Based on model confidence of the current delexicalized input, our algorithm improves delexicalization in every iteration to converge to the best input having the highest confidence. We show on benchmark and in-house datasets that our algorithm can greatly improve parsing performance for RNN based models, especially for out-of-distribution slot values.
\end{abstract}

%
%


\blfootnote{A version of this paper was published at INTERSPEECH 2019, Graz, Austria.} 

\section{Introduction} \label{sec:intro}

Spoken language understanding (SLU) models play a key role in modern voice controlled personal agents and AI chatbots. Given a user's utterance, a SLU model identifies the appropriate intent and slots from the utterance, which are subsequently used to fulfill the user command or continue dialog with the user. An intent classifier assigns the most likely intent label to the user utterance, while a slot tagger assigns a slot label to every word in the utterance. The slot values or informative words/phrases in the utterance are then extracted for executing the desired action. Correct identification of intent and slot values by SLU models play a vital role in the success of downstream tasks \cite{LiChenLiGaoC:17}. 

Recently, joint intent classification and slot labeling SLU models based on recurrent neural networks has been shown to achieve state-of-the-art performance on benchmark datasets \cite{HakTurCelChenGao:16,LiuLane:16,KimLeeStratos:17,GooGaoHsuHuo:18,WangShenJin:18}. However, these models often suffer from poor slot labeling accuracy when an utterance contain slots with large semantic variability, dissimilar to those encountered during training e.g. message text, partial show/artist names. In this paper, we refer such slots as {\em out-of-distribution (OOD) slots}. The main characteristics of such OOD slots are that they do not have fixed vocabulary, and moreover can take values with length and word distribution very dissimilar to their training vocabulary. Usually they also contain many out-of-vocab words. Standard SLU models are trained on expensive labeled training datasets where OOD slots with large semantic variability are never well represented. Presence of out-of-vocabulary words and phrases have been shown to further degrade performance of RNN based SLU models \cite{RayShenJin:18b}. As an example, in a Facebook dataset, a state-of-the-art bidirectional RNN based SLU model \cite{LiuLane:16} is particularly poor in identifying slots containing message text compared to other slot types as seen in Table \ref{tab:per_slot_facebook}.

A standard approach to tackle the high semantic variability of the input utterance is to preprocess the input, thereby replacing partly or wholly slot words/phrases with special tokens from training vocabulary, a process called delexicalization. For example, in Figure \ref{fig:delex_train} the word {\em ``Alice''} and phrase {\em ``happy birthday''} in the original utterance is replaced by special tokens $\langle contact \rangle$ and $\langle message \rangle$ respectively. The delexicalized utterance is then used by the model to infer the intent and slot labels of the original utterance. Delexicalization, mainly based on greedy longest string matching, has been explored before in SLU \cite{HecHak:12,HeckHakTur:13,lee2018coupled}. However, they do not always work well since they neglect any utterance context, and often resulting in errorneous input. Moreover, OOD slots with many out-of-vocab words (e.g. message text) are almost imposible to match and delexicalize. In \cite{ShinYooLee:18} the authors propose to improve slot filling in RNN models by using an additional delexicalized sentence generation task. Still it does not alleviate the problem of learning OOD slots since the overall task is again trained on the training data distribution. Using delexicalization to improve performance of other NLP systems has also been studied in the context of natural language generation \cite{JuraskaKarBowWalk:18}, dependency parsing \cite{zeman2008cross,DenDeh:17}, semantic parsing \cite{DongLap:16}, and representation learning \cite{lee2018coupled}.

\begin{table}[ht]
\caption{Slot labeling F1 score comparison for different slot categories in Facebook dataset using baseline Attention BiRNN parser \cite{LiuLane:16}.}\label{tab:per_slot_facebook}
\centering
\scriptsize
\setlength\tabcolsep{3pt}
\begin{tabular}{|l|c|c|c|}
\toprule
{\bf Slot category} & {\bf Total fraction} & {\bf Baseline F1 \%} & {\bf Our model F1 \%} \\
\midrule
Message text & $60.4\%$ & $85.16$ & $\mathbf{91.25}$ \\
Other non-message & $39.6\%$  & $96.8$ & $95.4$ \\ 
\midrule
Overall & $100\%$ & $89.82$ & $\mathbf{92.91}$ \\ 
\bottomrule
\end{tabular} 
\end{table}

In this paper, we develop a novel algorithm to iteratively delexicalize the input utterance guided by model's confidence on the current input as well as utterance context. Our approach allows effective delexicalization of OOD slot values, even when they have large semantic variability. Our delexicalization based hybrid parsing model is also modular and can be applied with any RNN based parser to improve its parsing performance. On both benchmark and in-house datasets, and using different RNN based parsers, our algorithm is demonstrated to significantly improve performance in both slot labeling and intent classification, achieving the state-of-the-art.

\section{Problem and background} \label{sec:problem}

In this section we formally define our spoken language understanding problem and the delexicalization approach to tackle it. Let a user provide an input utterance $\mathbf{x}=(x_1, \hdots, x_n),$ with words $x_i \in \mathcal{V}_T,$ the vocabulary. The task of an intent classifier is to determine the intent $I(\mathbf{x})$ of utterance $\mathbf{x},$ while a slot taggers labels each word $x_i$ with a slot label $y_i,$ thereby generating a sequence of slot labels $\mathbf{y} = (y_1, \hdots, y_n).$ A joint intent classifier and slot tagger model or parser $\mathcal{P}$ is jointly trained on the two tasks using a labeled training dataset $T.$ In delexicalization, we replace words/phrases in the input utterance $\mathbf{x}$ with some special tokens to generate a delexicalized input $\mathbf{x}' = (x_1', \hdots, x_m'),$ $m \leq n.$ A delexicalization algorithm performs this mapping $f: \mathbf{x} \to \mathbf{x}',$ possibly taking into account the training vocabulary $\mathcal{V}_T$ and current parse results $\{I(\mathbf{x}),\mathbf{y}\}.$  Now, the parser $\mathcal{P}$ is used to infer the slot labels $\mathbf{y}'$ and intent $I(\mathbf{x}')$ of the modified input $\mathbf{x}'.$ These results are subsequently used to infer (or modify) the parse of the original utterance $\mathbf{x}.$ Our iterative delexicalization algorithm performs this iteratively to converge at the best parsing result for the user utterance $\mathbf{x}.$ 

\paragraph{Recurrent neural network parsers:} RNN based parsers have been successfully applied for intent classification and slot labeling tasks in recent years \cite{MesDauYaoDengHakHe:15,HakTurCelChenGao:16,LiuLane:16,KimLeeStratos:17,GooGaoHsuHuo:18}. These models use a recurrent network (e.g. BiLSTM/GRU) to encode the input utterance $\mathbf{x}$ to a sequence of hidden state representations $\{\mathbf{h}_t\}_{t=1}^n.$ These hidden states, along with possible self-attention and/or intent attention are combined to generate the output slot label distribution $\{\mathbf{z}_t\}_{t=1}^n.$ Finally, the output slot labels are inferred as $y_t = \arg \max_{l_j} P(y_t=l_j) = \arg \max_{j} \mathbf{z}_t(j),$ where $l_j$ denotes the slot label corresponding to the $j-$th index.

\section{Our model} \label{sec:model}

In this section, we present our main iterative delexicalization algorithm. We refer as {\em slot words} the words which are informative of command parameters and has a ground-truth slot labels other than ``O''. {\em Context words} refer the words with slot label ``O'' which are either non-informative or conveys the utterance intent. Our hybrid SLU model can be viewed as a two step parser where the first step delexicalizes the input utterance $\mathbf{x}$ to $\mathbf{x}',$ and the second step uses a RNN based parser to infer the slot and intent labels $\{I(\mathbf{x}),\mathbf{y}\}$ using this delexicalized input $\mathbf{x}'.$ Our iterative inference algorithm executes these steps repeatedly until it converges to the best parse result. \par

\begin{figure}[hptb]
\begin{center}
\includegraphics[height =2.2in]{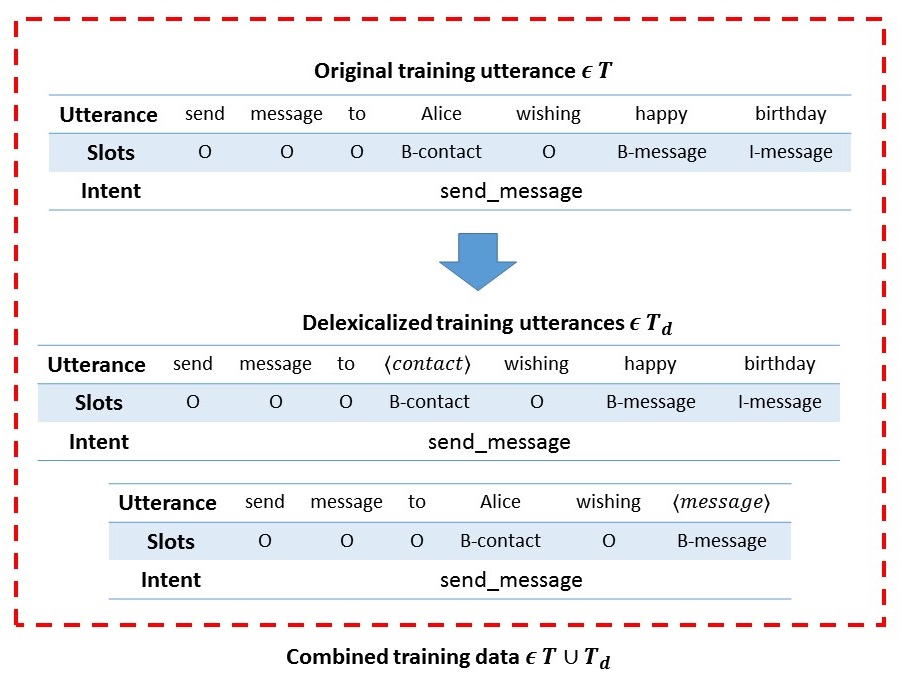}
\caption{Illustration of our combined dataset generation for training the iterative delexicalization parser $\mathcal{P}$.\label{fig:delex_train} }
\end{center}
\end{figure}

\paragraph{Idea:} The main idea behind our delexicalization based hybrid approach to parsing is based on the following observation. When an utterance is encountered having out-of-distribution slot words, the parser exhibits a low slot tagging confidence over these unknown words, often resulting in incorrect slot labels. However, if such slot words/phrases with low parsing confidence are replaced by suitable delexicalized tokens, which were also present in the training data, then typically there are two outcomes; {\bf (a)} parsing confidence on the new delexicalized utterance improves if delexicalization was performed correctly; or {\bf (b)} it degrades if delexicalization we performed incorrectly. Our iterative delexicalization algorithm exploits this phenomenon to generate a new set of candidate delexicalized utterances from the current input guided by parser's confidence in each iteration. Such iterations are continued as long as there exists at least one new candidate delexicalized input which improves parsing confidence over the previous set of candidate inputs. Finally, when there are no further confidence improvement the one having the best parsing confidence is used to generate the final parsing result.

\subsection{Iterative delexicalization algorithm}

We now describe our iterative delexicalization algorithm in details. The overall algorithm can be divided into two steps; first a model training step, and an iterative inference step. First we describe the model training step.

\paragraph{Model training:} To enable an RNN based parser to correctly parse a delexicalized input utterance it should correctly interpret the special delexicalized tokens from the utterance context. To enable this, the training set $T$ is augmented with a delexicalized training set $T_d$ where the ground truth slot words/phrases have been substituted by some special tokens, usually an unique token per slot type\footnote{An exception is slots which have a semantic hierarchy e.g. city name which can be both a source and destination slot. In such cases both slots can be replaced by the same token.}. However, to prevent the model to overfit to correlations between such special tokens, we substitute tokens randomly with probability $p_s<1$ (e.g. $p_s=0.75$). Then, the RNN parser $\mathcal{P}$ is trained on the combined training set $T' = T \cup T_d.$ In Figure \ref{fig:delex_train} we illustrate this training utterance generation in Facebook domain.   

\begin{figure}[hptb]
\begin{center}
\includegraphics[height =2.2in]{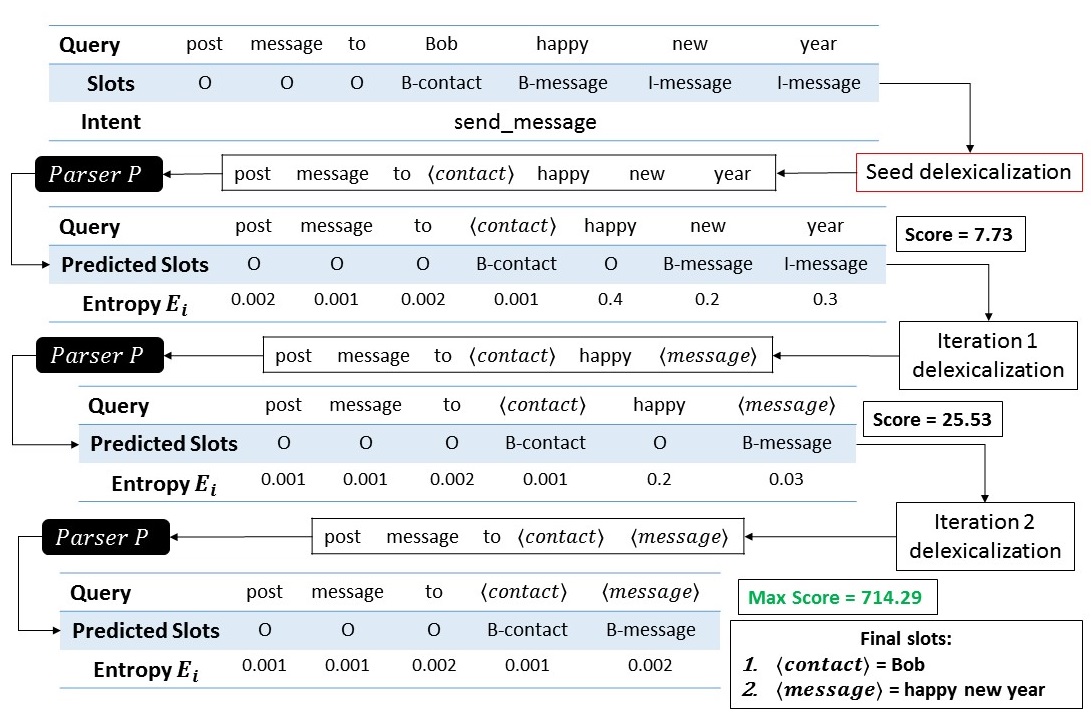}
\caption{Illustration of our iterative delexicalization inference Algorithm \ref{alg:iterative_delex}, using a threshold $\tau=0.05$. Note that iteration $1$ delexicalization is based on {\bf proper slot sequence}, and iteration $2$ delexicalization is based on {\bf special token expansion}.\label{fig:delex_infer} }
\end{center}
\end{figure}

\paragraph{Inference via iterative delexicalization:} When a user provides an input utterance $\mathbf{x},$ the iterative delexicalization Algorithm \ref{alg:iterative_delex} is used to infer the intent $I(\mathbf{x})$ and slot labels $\mathbf{y}$. At first the function $SeedDelexicalization(.)$ is used to generate an initial set $S$ of delexicalized utterances, each of which is parsed by $\mathcal{P}$ and given a parsing confidence using score function $Score(.)$. Next, in every iteration, each delexicalized input $\mathbf{x} \in S$ is used to generate more candidate delexicalized utterances using the function $Delexicalization(.).$ These new candidates are then parsed and their parsing scores are computed. We repeat this process till the maximum parsing confidence score no longer improves, or if no new candidates can be generated. Our algorithm also use a confidence threshold $\tau$ and a maximum size parameter $K$ to limit the amount of computation per iteration. We illustrate our iterative delexicalization inference algorithm using an example from Facebook domain in Figure \ref{fig:delex_infer}. Note that, our iterative algorithm is guaranteed to converge since each delexicalization step can only reduce the size of the input. Next we further elaborate the key subroutines and our score function.

\begin{algorithm}[ht]
\caption{Iterative delexicalization (inference)}
\label{alg:iterative_delex}
\begin{algorithmic}[1]
\Require Input utterance $\mathbf{x},$ parser $\mathcal{P},$ confidence threshold $\tau,$ size threshold $K$
\Ensure Intent $I(\mathbf{x}),$ slot labels $\mathbf{y}$

\State $S \gets SeedDelexicalization(\mathbf{x})$
\State $\text{maxConfidence} \gets 0$
\State Use $\mathcal{P}$ to parse each $\mathbf{x}' \in S.$ Compute confidence score $Score(\mathbf{x}')$ of each such parsing result
\State $\text{currentConfidence} \gets \max_{\mathbf{x}' \in S} Score(\mathbf{x}')$
\While{$\text{currentConfidence} > \text{maxConfidence}$ and $|S|>0$}
	\State $S' \gets S,$ and $S \gets \emptyset$
	\State If $|S'| > K,$ pick top $K$ delexicalized utterance $\mathbf{x}'$ with highest confidence score $Score(\mathbf{x}')$ in previous iteration
	\For{$\mathbf{x}' \in S'$}
		\State $S \gets S \cup Delexicalization(\mathbf{x}', \mathcal{P}, \tau)$
	\EndFor
	\State Use $\mathcal{P}$ to parse each $\mathbf{x}' \in S.$ Compute confidence score $Score(\mathbf{x}')$ of each such parsing result
	\State $\text{currentConfidence} \gets \max_{\mathbf{x}' \in S} Score(\mathbf{x}')$
\EndWhile
\State $\mathbf{x}_{best} \gets \arg \max_{\mathbf{x}' \in S'} Score(\mathbf{x}')$
\State Infer intent $I(\mathbf{x}),$ slot labels $\mathbf{y}$ using the parsing results $I(\mathbf{x}_{best}),\mathbf{y}_{best}$ of $\mathbf{x}_{best}$
\State Output intent $I(\mathbf{x}),$ slot labels $\mathbf{y}$
\end{algorithmic}
\end{algorithm} 

\paragraph{Seed delexicalization:} To generate the starting set of delexicalized input utterances we use a $SeedDelexicalization(.)$ subroutine. This function uses longest first string matching to identify possible word/phrase that can be delexicalized. However, unlike replacing all matched substrings greedily similar to \cite{HecHak:12,HeckHakTur:13}, our algorithm generates two candidates per match, one with and one without delexicalizing the matched substring. This ensures that if a substring is matched incorrectly (i.e. it also includes context words) at least one of the candidate is still correct (the one not delexicalized). We also do not match words/phrases which appeared as context words/phrases in the training data, and slot words/phrases which may be part of two different slot tags (e.g. an artist who is both actor and director slot). However, this step is not expected to delexicalize OOD slots such as message text.

\paragraph{Model based delexicalization:} In the main iteration loop we delexicalize using a subroutine $Delexicalization(.).$ Unlike the previous $SeedDelexicalization$ function which uses string matching, the $Delexicalization$ function performs delexicalization based on parsing output of $\mathbf{x}'$ and some domain independent rules as described next. Recall that the RNN based parser $\mathcal{P}$ generates the slot label distribution $\mathbf{z}_t$ for every word/token $x_t$ in the input. We use the entropy function $E_t = Entropy(\mathbf{z}_t)$ to first capture the parser's slot level confidence for the $t$-th word. The main motivation behind the rules in $Delexicalization$ function is to refine an incorrect delexicalized candidate input towards a better candidate.   
\begin{enumerate}[leftmargin=*]
\item {\em Proper slot sequence:} Suppose slot words/phrases have been partly or wholly identified by the model using a sequence of tags $(B\mhyphen Y,I\mhyphen Y,\hdots,I\mhyphen Y)$ for a slot tag $Y,$ and which already do not contain any special tokens. Then these are replaced by the special token corresponding to slot tag $Y,$ to generate the delexicalized candidate for next iteration. An example application of this rule is demonstrated in Figure \ref{fig:delex_infer}.

\item {\em Improper slot sequence:} Often for OOD slots the model may improperly identify the slot words and assign a sequence of labels $(I\mhyphen Y,\hdots,I\mhyphen Y)$ without any beginning token $B\mhyphen Y.$  Then, such sequences are also replaced by the special token corresponding to slot tag $Y,$ to generate the next delexicalized candidate.

\item {\em Special token expansion:} For an out-of-distribution slot value of large size (e.g. a long message text), the parser may correctly identify only subset of the slot words. Applying the previous rules only delexicalizes part of this slot value. Hence such partial delexicalization needs to be expanded to include the remaining slot words. This is performed as follows. Starting with a special token $\langle Y \rangle$ at position $t,$ any contiguous sequence of words $(x_s, \hdots, x_{t-1})$ and $(x_{t+1}, \hdots , x_{r})$ before/after token $\langle Y \rangle,$ which satisfies the condition $E_i > \tau,$ for $i \in [s,t-1]\cup [t+1,r]$ and $E_{s-1}\leq \tau,$ $E_{r+1} \leq \tau,$ are also delexicalized by the same token $\langle Y \rangle$ to generate a new candidate.
\end{enumerate}

For each candidate in current iteration, we apply all three rules to generate candidates for next iteration. We also ignore duplicate candidates.

\paragraph{Confidence score:} To evaluate the confidence of $\mathcal{P}$ on the current parsing output for an input $\mathbf{x}',$ we use a score function $Score(\mathbf{x}')$ in Algorithm \ref{alg:iterative_delex}. We implement this score function as the inverse average entropy of slot label distribution in the parser output. More precisely we define:
\begin{equation}
Score(\mathbf{x}') = n/\sum_{t=1}^n E_t = n/\sum_{t=1}^n Entropy(\mathbf{z}_t) \label{eq:score}
\end{equation}

We note that, the iterative delexicalization step is performed mainly to improve the delexicalization of OOD slots and may not significantly improve slot labeling performance of other slots. Hence in our experiments we perform the seed delexicalization step for every slot, but restrict the iterative delexicalization step only to specific OOD slots. One may be concerned that the iterative inference of our algorithm can be slow. However, in each iteration the delexicalized candidates can be parsed independently in parallel using batch inference. We also observe our algorithm to converge fast within $2$ or $3$ iterations in benchmark and in-house datasets, even with long message text slots. The average inference time observed was around $0.08$ seconds, fast enough for any practical systems.  

\paragraph{Discussion:} An alternative approach to improve the performance of RNN parsers is to add an output CRF layer to optimize over label sequence via dynamic programming \cite{YaoPengZweYuLiGao:14}. Our algorithm can also be viewed as performing optimization over the space of delexicalized inputs using dynamic programming. In experiments (Section \ref{sec:experiment}) we show that our model even outperform baseline RNN parsers with output CRF layer. Beam search can also generate multiple slot label sequences. However, using our score metric in equation \ref{eq:score}, all such beam search candidates would receive the same score. Since our delexicalization approach generates candidates with different input lengths and tokens, they can be sufficiently differentiated using our score function.  Our proposed system also has the added advantage that during training time, for a slot with very large vocabulary (e.g. movie/show titles), it is sufficient to just train with few delexicalized utterances instead of using examples from the entire vocabulary, thereby reducing training time considerably. During inference the seed delexicalization step ensures that such instances will be delexicalized and correctly labeled by the model. Many slot values (e.g. artist, song titles) also experience a distribution shift over time, and may require sophisticated models and retraining to adapt to such changes \cite{KimStraKim:17}. Our model can easily support such domain adaptation/continuous learning without any model retraining by just updating training vocabulary with contemporary slot values. 

\section{Experiments} \label{sec:experiment}

In this section we present our main experimental results.

We evaluate our algorithm on two different datasets. First, we use the benchmark SNIPS dataset \cite{SNIPS:18}, and second an in-house Facebook dataset. The SNIPS dataset has total \numprint{13784} training and \numprint{700} test utterances. Our Facebook dataset contains a total of \numprint{1191} utterances. SNIPS dataset contain OOD slots such as $object\_name,$ $movie\_name;$ where as Facebook dataset contain OOD slots such as $message.$ Due to the smaller size, in Facebook domain we perform a $5$ fold validation using five different splits and averaging the result across split. Our datasets are summarized in Table \ref{tab:datasets}. Facebook dataset has a larger out-of-vocab percentage due to presence of OOD message text slot. We do not consider the popular ATIS dataset since it lacks OOD slots, hence not suited for our evaluation. As baseline, we use two state-of-the-art RNN based parsers which perform joint intent classification and slot labeling. First, the attention based BiRNN parser by Liu and Lane \cite{LiuLane:16}, second is the recent Slot Gated parser by Goo et al. \cite{GooGaoHsuHuo:18}. We combine our iterative delexicalization algorithm using both the baseline parsers as $\mathcal{P},$ and compare their performance on both the datasets. We also implement few advanced baselines. First we augment an output CRF layer to attention BiRNN parser to test the effect of output sequence level optimization for OOD slots. We also show the performance of our models when delexicalization is only performed using greedy longest substring matching \cite{HecHak:12,HeckHakTur:13}. Our evaluation metric is both intent classification accuracy, and slot labeling F1 score.

For baseline models we use code made available by the authors of \cite{LiuLane:16,GooGaoHsuHuo:18}. We train the baseline models using their default parameters. To integrate the baseline models as base parser $\mathcal{P}$ with our iterative delexicalization algorithm, we also train them on the combined training set $T'$ using the same default parameters. We use a slot confidence threshold $\tau=1\times 10^{-5}.$ 

\begin{table}[ht]
\caption{Statistics of datasets used in our experiments.}\label{tab:datasets}
\centering
\small
\setlength\tabcolsep{4pt}
\begin{tabular}{|l|c|c|c|c|c|c|}
\toprule
{\bf Dataset} & {\bf Total size} & {\bf \#Splits} & {\bf \#Intents} & {\bf \#Slots} & {\bf \#Vocab} & {\bf \#OOV} \\
\midrule
SNIPS & \numprint{14484} & $1$ & $7$ & $52$ & \numprint{11604} & $2.9$ \% \\
Facebook & \numprint{1191} & $5$ & $9$ & $4$ & \numprint{1158} & $4.7$ \% \\ 
\bottomrule
\end{tabular} 
\end{table} 

\paragraph{Results:} First, we compare the performance of all models on the benchmark SNIPS dataset. Table \ref{tab:perf_snips} presents the numerical results. We observe that for both the RNN based parsers our model using iterative delexicalization achieves both highest slot tagging F1 score and intent accuracy. While adding a CRF layer to baseline RNN parser improves slot labeling performance, it still perform worse than our algorithm. Moreover, our algorithm also performs better than greedy longest substring matching based delexicalization. This is because string matching based delexicalization can often result in erroneous match, and cannot match out-of-vocab words/phrases. Using attention BiRNN parser and iterative delexicalization, our model also achieves state-of-the-art slot labeling F1 score of $\mathbf{93.24}\%$ and intent accuracy $\mathbf{98.57}\%.$ Our baseline Slot Gated parser perform slightly worse on SNIPS than \cite{GooGaoHsuHuo:18} since we use default parameter settings without parameter tuning. In Table \ref{tab:perf_facebook}, we compare the performance of all models on Facebook dataset. Once again we observe that our iterative delexicalization algorithm achieves significant improvement of slot tagging F1 score as well as intent accuracy. The gain is particularly high due to the presence of OOD message slots in Facebook, which show a large semantic variability in the test set.  

\begin{table}[ht]
\caption{Intent classification and slot labeling performance comparison of all models on benchmark SNIPS dataset.}\label{tab:perf_snips}
\centering
\scriptsize
\setlength\tabcolsep{2pt}
\begin{tabular}{|l|c|c|}
\toprule
{\bf RNN Parser} & {\bf Slot tagging F1 \%} & {\bf Intent accuracy \%} \\
\midrule

Slot Gated BiRNN \cite{GooGaoHsuHuo:18} & $88.80$ & $97.00$ \\
Capsule NLU \cite{ZhaLiDuFanYu:18} & $91.80$ & $97.30$ \\ 
\midrule
Attention BiRNN (our baseline) & $90.64$ & $98.00$ \\
Attention BiRNN + CRF & $91.91$ & $98.00$ \\
Attention BiRNN + greedy delex.  & $92.56$ & $98.29$ \\
Attention BiRNN + iterative delex. & $\mathbf{93.24}$ & $\mathbf{98.57}$ \\ 
\midrule
Slot Gated BiRNN (our baseline) & $85.25$ & $93.14$ \\
Slot Gated BiRNN + greedy delex. & $86.83$ & $94.86$ \\
Slot Gated BiRNN + iterative delex. & $\mathbf{88.14}$ & $\mathbf{95.14}$ \\ 
\bottomrule
\end{tabular} 
\end{table}

We also perform some slot level error analysis to observe the advantage of our iterative delexicalization algorithm for out-of-distribution slots. For Facebook dataset, we divide the slots into two categories; first those of the form of {\em message text}, and the {\em other non-message} slots, since message texts can often have large semantic variability. Using the Attention BiRNN parser \cite{LiuLane:16} as baseline, we plot the average F1 score for a particular train-test split for these two slot categories in Table \ref{tab:per_slot_facebook}. We observe that our iterative delexicalization algorithm significantly improves the F1 score of message text slots while maintaining similar F1 score for other slots.

\begin{table}[ht]
\caption{Performance comparison of all models on Facebook dataset. We compute $5$ fold average performance in this dataset due to smaller size.}\label{tab:perf_facebook}
\centering
\scriptsize
\setlength\tabcolsep{2pt}
\begin{tabular}{|l|c|c|}
\toprule
{\bf RNN Parser} & {\bf Slot tagging F1 \%} & {\bf Intent accuracy \%} \\
\midrule
Attention BiRNN (our baseline) & $84.46$ & $93.80$ \\
Attention BiRNN + CRF & $86.79$ & $93.80$ \\
Attention BiRNN + greedy delex. & $85.65$ & $94.31$ \\
Attention BiRNN + iterative delex. & $\mathbf{89.22}$ & $\mathbf{94.82}$ \\ 
\midrule 
Slot Gated BiRNN (our baseline) & $84.86$ & $93.46$ \\
Slot Gated BiRNN + greedy delex. & $86.49$ & $92.72$ \\
Slot Gated BiRNN + iterative delex. & $\mathbf{88.89}$ & $\mathbf{93.48}$ \\ 
\bottomrule
\end{tabular} 
\end{table}

\section{Conclusion} \label{sec:conclusion}

State-of-the-art RNN based joint SLU models perform poorly on certain out-of-distribution slots which may encounter slot values with large semantic variability after deployment. Previous string matching based delexicalization techniques are inadequate to handle such slots since these slot values are mostly out-of-vocab. In this paper we propose a novel iterative delexicalization algorithm which exploits model uncertainty to improve delexicalization for such out-of-distribution slots. In addition, our model also enables faster model training for slots with large training vocabulary (e.g. movie/show titles), and supports some continuous learning without requiring model updates by simply maintaining an updated contemporary vocabulary. In experiments on benchmark and in-house datasets, we demonstrate significant improvement in SLU performance using our algorithm, thereby achieving state-of-the-art results.

%

\bibliographystyle{abbrvnat}
\bibliography{delex}

%
%

\end{document}